\pdfoutput=1

\documentclass[11pt]{article}

\usepackage[final]{acl}

\usepackage{times}
\usepackage{latexsym}
\usepackage{amsmath}

\usepackage[T1]{fontenc}

\usepackage[utf8]{inputenc}

\usepackage{microtype}
\usepackage{microtype}
\usepackage{float}

\usepackage{inconsolata}

\usepackage{graphicx}
\usepackage{makecell}
\usepackage{booktabs}
\usepackage{multirow}
\usepackage[dvipsnames]{xcolor}

\newcommand{\ul}[1]{\underline{#1}}

\newif\ifhalueval
\haluevalfalse  

\newcommand{\ifhalu}[2]{\ifhalueval #1\else #2\fi}

\newif\ifshort
\shorttrue  

\newcommand{\is}[2]{\ifshort #1\else #2\fi}

\title{IKnow:
Instruction-Knowledge-Aware Continual Pretraining\\for Effective Domain Adaptation}

\author{Tianyi Zhang\thanks{Equal contribution.} \and Florian Mai\footnotemark[1] \and Lucie Flek\thanks{Corresponding author.}  \\
University of Bonn\\
Lamarr Institute for Machine Learning and Artificial Intelligence\\
\texttt{tianyiz0423@gmail.com} \\ \texttt{fmai@uni-bonn.de} \\ \texttt{flek@bit.uni-bonn.de}}

\begin{document}
\maketitle
\begin{abstract}
\emph{Continual pretraining} promises to adapt large language models (LLMs) to new domains using only unlabeled test-time data, but naively applying standard self-supervised objectives to instruction-tuned models is known to degrade their instruction-following capability and semantic representations. 
Existing fixes assume access to the original \textit{base} model or rely on knowledge from an \textit{external} domain-specific database - both of which pose a realistic barrier in settings where the base model weights are withheld for safety reasons or reliable external corpora are unavailable.
In this work, we propose \textbf{Instruction-Knowledge-Aware Continual Adaptation (IKnow)}, a simple and general framework that \emph{formulates novel self-supervised objectives in the instruction-response dialogue format}. Rather than depending on external resources, IKnow leverages domain knowledge embedded in the text itself and learns to encode it at a deeper semantic level. 
\is{}{Specifically, we propose (i) 
\textbf{Masked Phrase Prediction} (MPP), where semantically meaningful phrases are masked and reconstructed, and (ii) \textbf{NL–KG Loop Prediction}, where the model is trained to perform bidirectional transformations between natural language and knowledge-graph tuples.}
\is{}{Comprehensive experiments on knowledge-intensive downstream tasks show that IKnow enables improved domain-specific adaptation for question answering
without requiring access to restricted base models or external knowledge-bases.}
\end{abstract}

\section{Introduction}
Large language models (LLMs) that have undergone \emph{scaled pretraining} and further \emph{instruction tuning}~\citep{wei2022finetuned} now underpin most modern NLP applications.  
Although these models achieve strong zero-shot performance, their accuracy degrades when the deployment domain drifts from the public web data on which they were pre-trained, i.e., when asked questions about world events that happened after the curation of the pretraining data.
A classical remedy is \emph{continued pre-training} on unlabelled in-domain text, first popularized by \citet{gururangan-etal-2020-dont} for BERT-style~\citep{devlin2019bert} models. However, continued prerain models naively with a next-token-prediction \is{}{ (NTP) or a masked-language-model (MLM) }objective can encounter undesired outcomes in instruction-tuned models, such as catastrophic forggetting~\citep{jindal2024balancing}.

Firstly, the widely used instruction-tuned models can lose their instruction-following ability thereafter.
\citet{fleshman2024readapt} mitigate this effect by performing continual finetuning on the \emph{base} model and adding the instruction-following ability back in by adding an instruction-task vector to the fine-tuned model weights. However, oftentimes LLM developers do not publish the base model to avoid safety risks that may emerge from models without post-training, e.g. Phi-4~\citep{abdin2024phi}, making this approach unviable.

Secondly, knowledge-intensive downstream tasks, such as question answering \ifhalu{ and summarization,}{}
require a deeper encoding of semantics rather than shallow sequence probabilities. To facilitate knowledge encoding, \citet{zhang-etal-2019-ernie, peters-etal-2019-knowledge, wang-etal-2021-kepler} inject the knowledge embeddings acquired from external dataset into token embeddings in pre-training. However, this approach is unviable in continual pretraining phase due to the acquisition of compatible knowledge embeddings and modifications in the model architectures.

To address domain drift declination in an off-the-shelf instruction-tuned model scenario, we introduce \textbf{Instruction-Knowledge-Aware Continual Adaptation} (\textsc{IKnow}), a light-weight continual pretraining framework that perform knowledge-aware self-supervised objectives in the format of an instruction–response dialogue to encourage the model to keep its instruction following ability.
completion. 
Additionally, we propose two knowledge-aware pretraining tasks to enhance a model's ability to capture entity-relation knowledge
\is{
by constructing relations from parse trees. 
}
{. In the Masked Phrase Prediction task, the model is trained to predict phrases of entities, as identified by constituency parsing, rather than randomly selected tokens. In the NL-KG Loop Prediction task, the model is guided to perform bidirectional translation between natural language and knowledge graphs extracted from dependency parsing, thereby encouraging a deeper integration of entity-relation structure into the model representations.}
Our experimental evaluation on two question answering tasks demonstrates the effectiveness of our approach.

\begin{figure*}[ht]
  \centering
  \includegraphics[width=0.95\textwidth]{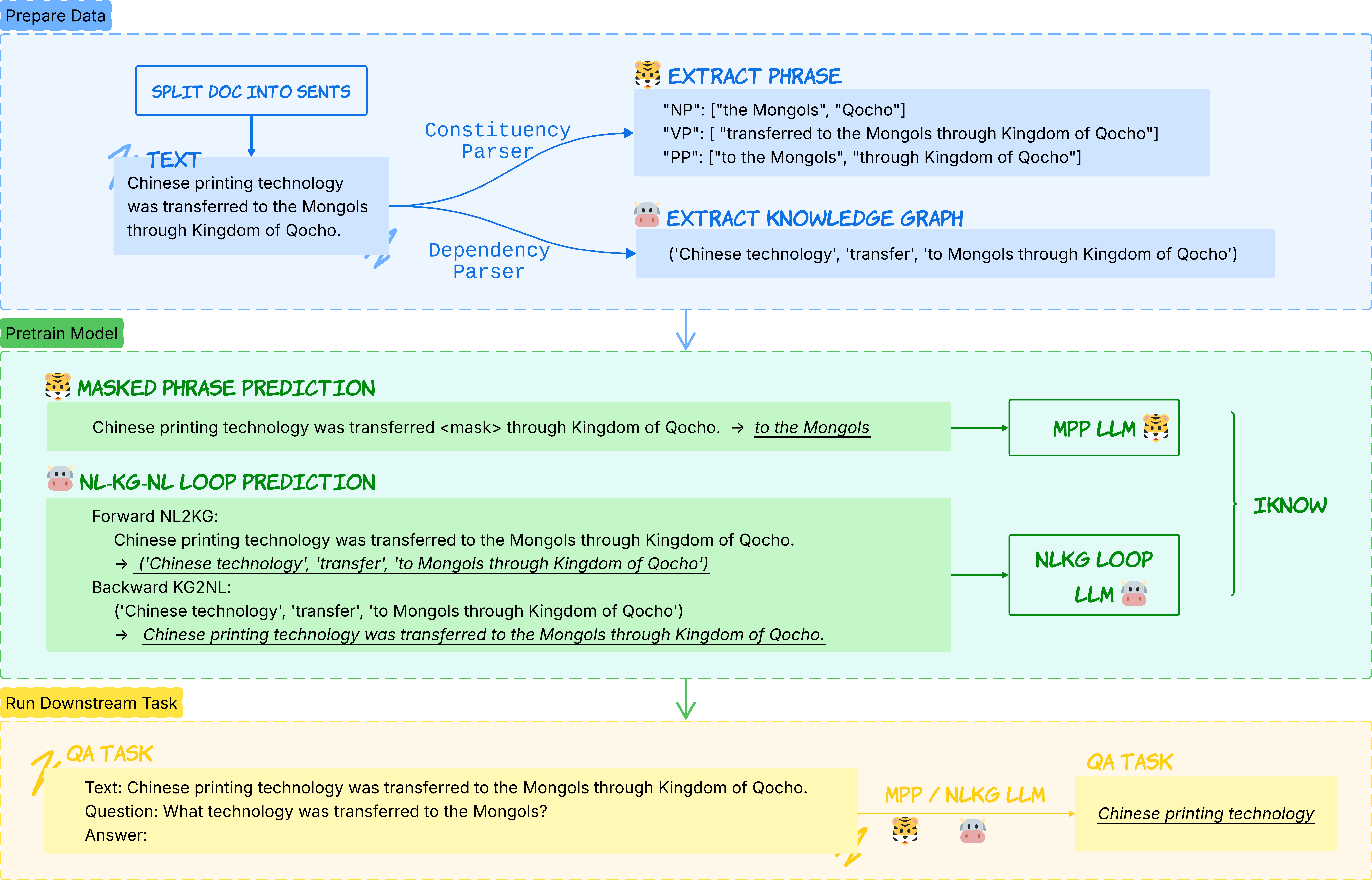}
  \caption{
  Overview of our framework, consisting of three stages: \textbf{(I)} \textcolor{RoyalBlue}{data preparation}, \textbf{(II)} \textcolor{OliveGreen}{model pretraining}, and \textbf{(III)} \textcolor{BurntOrange}{downstream evaluation}.
\textbf{(I)} In the data preparation stage, we leverage off-the-shelf syntactic parsers to extract structural information from text. Specifically, we employ two complementary strategies: (1) extracting constituency to identify phrases, and (2) deriving knowledge graphs based on dependency structures.
\textbf{(II)} In the pretraining stage, we introduce two distinct objectives: \textbf{Masked Phrase Prediction (MPP)}, which trains the model to reconstruct syntactically meaningful spans based on constituency structure, and the \textbf{NL-KG Loop} task, which encourages bidirectional reasoning between natural language and structured knowledge graphs with a forward pass NL2KG and a backward pass KG2NL.
\textbf{(III)} Finally, we evaluate the resulting models—\textbf{MPP-LLM} and \textbf{NLKG-Loop-LLM}—on knowledge-intensive downstream tasks such as question answering, demonstrating the effectiveness of structure and semantic informed pretraining.
All our tasks are knowledge-aware and formatted in the \emph{instruction-response} template. The response is in \underline{\textit{underlined italic format}} following an arrow.}
  \label{img:system}
\end{figure*}

\section{Related Work}
\label{sec:related}

\paragraph{Continual pretraining.}
\citet{gururangan-etal-2020-dont} first showed that running extra unsupervised epochs on task-specific corpora (``domain–adaptive'' or ``task–adaptive'' pre-training) boosts downstream accuracy.
Subsequent studies asked whether the same recipe helps \emph{instruction-tuned} checkpoints: \citet{jindal2024balancing} report severe alignment loss, while \citet{fleshman2024readapt} recover alignment by re-injecting the base–to–instruction weight delta—an option unavailable when the base model is unreleased.
Our work stays in the continual-pretraining paradigm but replaces the vanilla MLM loss with an \textit{instruction-style} variant that keeps the dialogue context intact.

\paragraph{Test-time training and adaptation.}
Test-Time Training (TTT) updates model parameters on \emph{each individual} test example, usually with a self-supervised loss \citep{sun2020test}; Test-Time Adaptation extends this idea to streams of test batches \citep{wang2021tent}.
Early NLP instances include T-SAS for QA \citep{jeong2023tsas} and few-shot experiments by \citet{akyurek2024surprising}.  
Although we pretrain on the entire test corpus \textit{before} questions arrive—hence diverging from the per-sample TTT setting—the two lines of work are complementary, and we plan to explore a true TTT variant of our objective in future work. 

\paragraph{Knowledge-intensive pretraining.}
Embedding strategies that explicitly target \emph{knowledge} tokens yield larger downstream gains than vanilla MLM.  
\citet{zhang-etal-2019-ernie} concatenate the entity embeddings with token embeddings. \citet{peters-etal-2019-knowledge} inject the external database, e.g. ConceptNet, knowledge embeddings into the original token embeddings. \citet{wang-etal-2021-kepler} combines entity de1scription embeddings with token embeddings. Another line of work encoding semantics through masked prediction on meaningful tokens.
\citet{joshi-etal-2020-spanbert} first demonstrated that masking entire \textit{spans} boosts performance on question answering tasks.  
Building on that idea, \citet{golchin-etal-2023-mask} show that during continued pretraining it is more efficient to mask in-domain keywords, while \citet{kohli2025choose} introduce a curriculum that gradually shifts the mask toward domain-specific concepts, cutting compute during biomedical adaptation by an order of magnitude.
Our work aligns with the previous research on train the model with semantic knowledge, but we focus on the knowledge embedded in the text itself, not requiring any external dataset.

\section{Method}
\label{sec:method}

\subsection{Problem Setup}
We assume a standard extractive QA benchmark\is{}{ such as \textsc{SQuAD}}, whose test portion
is a collection of triples
\(
\mathcal{T}
=\{(C_i,\,Q_i,\,A_i)\}_{i=1}^{N},
\)
where each \emph{context}~$C_i$ is a supportive context document, $Q_i$ is a natural–language question with answer
\(A_i\).
In many real deployments the \emph{passages become available long before the questions are
asked}.  
A concrete example is an enterprise assistant that must answer
employee queries about a newly published internal manual: the manual (our~$C_i$) can be processed
offline, whereas the questions~$Q_i$ only arrive during usage.
Hence, we assume access to the contexts of the test set.

Formally, at time~$t{=}0$ we receive the unlabeled \emph{context set}
\(\mathcal{C}_{\text{test}}=\{C_i\}_{i=1}^{N}\).
We are given an LLM with parameters
\(\boldsymbol\theta_0\),
but we \emph{do not} have access to the underlying base checkpoint.
Our goal is to adapt~$\boldsymbol\theta_0$ using only~$\mathcal{C}_{\text{test}}$,
then answer all $(Q_i)$ at $t{>}0$.
Crucially, our LLM with parameter $\theta_0$ is \emph{instruction-tuned}, posing the challenge how to retain its instruction-following ability during continual pretraining.

\subsection{Instruction-Knowledge-Aware Continual Adaptation (IKnow)}

Instruction-tuned methods are trained on input data that usually have the following format: 
\begin{align*}
\texttt{|user| <user\_query>} \\
\texttt{|assistant| <response>},
\end{align*}
where 
\texttt{<user\_query>} is the instruction to be followed, and \texttt{<response>} is the reply from the assistant. Note that the LLM is only trained on the response.

In order to retain the instruction-following ability of instruction-tuned LLMs during continual pretraining, our main idea is to transform the context set $\mathcal{C}_{\text{test}}$ into $(user\_query, response)$ pairs. Figure~\ref{img:system} summarizes our approach.

\paragraph{Prepare data}
In order to make best use of the often limited data we have available, we first split the context $c$ into sentences $S_c = \{ s | s \text{ is sentence in } c\}$.
Then we use a constituency parser to identify phrases $P_s = \{p_i | p_i \in const(s)\}$, and a dependency parser to identify the set of $(subject, root, object)$ relations in s, $KG_{s} = \{(s, r, v) \in dep(s)\}$.

After parsing the data, we generate instruction-tuning training examples in three different ways: Masked Token Prediction (MTP), Masked Phrase Prediction (MPP), and NL$\leftrightarrow$KG.

\paragraph{Masked Token Prediction}
Given a (tokenized) sentence \( s = s_1 \dots s_n \), we randomly choose an index \( 1 \leq i \leq n \) to mask out. We then set

\begin{quote}
\texttt{\textless user\_query\textgreater} = \texttt{Complete the masked token:} $s_1 \dots s_{i-1}$ \texttt{\textless mask\textgreater} $s_{i+1} \dots s_n$\\
\texttt{\textless response\textgreater} = $s_i$
\end{quote}

MTP can be considered a straigth up adaptation of standard masked language modeling~\citep{devlin2019bert} to the instruction-tuning case.

\paragraph{Masked Phrase Prediction}

Adaptation to new domains requires an understanding of the relevant entities and their relations in that domain, especially for knowledge intensive tasks. While MLM is a good option for teaching general language understanding ability, it chooses masked tokens randomly rather than focusing on entities as humans. To address this, we propose to mask out an entire phrase, e.g. noun phrase, verb phrase, or prepositional phrase, which correspond more to entities (NPs and PPs) or relations (VPs).

Analogous to before, we randomly select a phrase $p_j = s_k \dots s_{k+l}$ of length $l$ that will be masked out:

\begin{quote}
\texttt{\textless user\_query\textgreater} = \texttt{Complete the masked words:} $s_1 \dots s_{i-1}$ \texttt{\textless mask\textgreater} $s_{i+1} \dots s_n$\\
\texttt{\textless response\textgreater} = $p_j$
\end{quote}

\paragraph{NL$\leftrightarrow$KG}

To emulate the human learning process—where input is encoded into structured knowledge and output is decoded from it \cite{ATKINSON196889}—we propose a framework that explicitly constructs a knowledge graph from a natural language sentence ($s$), referred to as NL$\rightarrow$KG. We formulate the task as follows:
\begin{quote}
\texttt{<user\_query> = Please extract knowledge tuples (subject, verb, object) from the text: s \\ <response> = $KG_s$}
\end{quote}
The KG$\rightarrow$NL task asks for the reverse.




\section{Experiments}

\subsection{Experimental Setup}

\paragraph{Hypotheses and baselines}
Our experiments are designed to test two hypotheses: (\textbf{H1}) Instruction-style pretraining tasks improves over the naive next-token prediction (\textbf{NTP}) baseline. (\textbf{H2}) Pretraining tasks tailored toward knowledge-intensive tasks, namely Masked Phrase Prediction (\textbf{MPP}) and NLKG, improve performance compared to naive Masked Token Prediction (\textbf{MTP}).

\paragraph{Datasets} 

To demonstrate the effectiveness of our approach on two recently published, knowledge-intensive question answering datasets.
\ifhalu{HaluEval \cite{li2023halueval} consists of documents and summaries collected from multiple sources, with hallucinated summaries generated by manually crafted model outputs. 
HaluSum and HaluNLI are modifications of this dataset by \citet{anonymous_halunli}. Given the original dataset containing triplets of \textit{(document, reference summary, hallucinated summary)}, we construct two tasks: 
For the HaluSum task, we use the \textit{(document, reference summary)} pairs to assess the summarization. For the HaluNLI task, we frame \textit{document} as the premise and \textit{reference summary or hallucinated summary} as the hypothesis to formulate an entailment vs.\ contradiction classification problem. }{}RepliQA \cite{monteiro2024repliqa} includes human-curated news articles paired with automatically generated question-answer pairs. 
SciQAG \cite{wan2024sciqag} focuses on scientific publications with corresponding generated QA pairs. 
Detailed statistics for each dataset are provided in Appendix \ref{app:dataset}.
We adopt different evaluation metrics as provided by \href{https://huggingface.co/docs/evaluate/en/index}{Hugging Face Evaluation package} based on the characteristics of the datasets. On \ifhalu{HaluSum, and }{}RepliQA and SciQAG, which contain long-form answers, we report ROUGE-L-F1. 
\ifhalu{On HaluNLI we use accuracy.}{}

\paragraph{Hyperparameters}
To demonstrate the generalizability of our results, we perform our experiments with two models (\href{https://huggingface.co/meta-llama/Llama-3.2-3B-Instruct}{Llama-3.2-3B-Instruct} and \href{https://huggingface.co/Qwen/Qwen3-1.7B}{Qwen3-1.7B}) and two fine-tuning techniques, full-finetuning and LoRA~\citep{hu2022lora}.
Learning rates are tuned per model and fine-tuning mode. The full description of the experimental setup can be found in Appendix~\ref{sec:experimental-setup}.

\subsection{Results}\label{sec:results}

\ifhalueval
\begin{table}
    \centering
    \resizebox{0.48\textwidth}{!}{%
    \begin{tabular}{c|c|c|c}
    \toprule
         & HaluSum & RepliQA & SciQAG \\ \midrule
         \multicolumn{4}{c}{\textit{Llama-3.2-3B-Instruct}} \\
         \hline
         Llama-3.2-3B-Instruct & 22.98 & 34.71 & 39.52 \\
        \hline
         NTP (no instruct)  & &  &  39.41\\
         \hline
         MTP (instruct)  & 22.43 & 35.47 &  39.66\\
         MPP (instruct)  & \textbf{23.02} & 35.58 &  40.11\\
         NLKG (instruct) & 23.11 & \textbf{36.83} &  39.88\\ \midrule
         \multicolumn{4}{c}{\textit{Qwen3-1.7B}} \\
         \hline
         Qwen3-1.7B & 18.85 & 28.73 & 37.54 \\
         \hline
         NTP  (no instruct)      & 19.02 & 28.8  & 37.10 \\
         \hline
         MTP (instruct)       & 18.60 & \textbf{31.71} & \textbf{38.55} \\
         MPP  (instruct)      & 18.76 & 30.42 & 37.94 \\
         NLKG (instruct)      & \textbf{19.55} & 29.4 & 38.30 \\
    \bottomrule
    \end{tabular}
    }
    \caption{Results \textbf{with full fine-tuning}.}
    \label{tab:results-nolora}
\end{table}
\else
\begin{table}[t]
    \centering
    \resizebox{0.4\textwidth}{!}{%
    \begin{tabular}{c|c|c}
    \toprule
        Model & RepliQA & SciQAG \\ \midrule
         \hline
         \textbf{Llama-3.2-3B-Instruct} & 34.71 & 39.52 \\  \midrule
 \multicolumn{3}{c}{Full Parameter}\\
        \hline
         NTP  &  0.2* &  39.41\\
         MTP  & \ul{35.47}&  \ul{39.66}\\
         MPP  & \it{\ul{35.58}}&  \it{\ul{\textbf{40.11}}}\\
         NLKG & \it{\ul{\textbf{36.83}}}&  \it{\ul{40.00}}\\ \midrule
 \multicolumn{3}{c}{LoRA}\\     \hline
         NTP & 37.19 &  39.89\\
         MTP & 28.13 & 39.24 \\
         MPP & \it{\ul{\textbf{38.62}}} &  39.12\\
         NLKG &  \it{34.32} &  \it{39.26}\\ 
         \midrule
         \hline
         \textbf{Qwen3-1.7B} & 28.73 & 37.54 \\  \midrule
 \multicolumn{3}{c}{Full Parameter}\\
         \hline
         NTP  & 28.8  & 37.10 \\
         MTP  & \ul{\textbf{31.71}} & \ul{\textbf{38.55}} \\
         MPP  & \ul{30.42} & \ul{37.94} \\
         NLKG & \ul{29.4} & \ul{38.30} \\  \midrule
 \multicolumn{3}{c}{LoRA}\\
                   \hline
         NTP & 29.31 & 37.78 \\
         MTP & \ul{\textbf{31.86}} & \ul{38.56} \\
         MPP & \ul{31.39} & \it{\ul{\textbf{38.96}}} \\
         NLKG & \ul{29.53} & \ul{38.55} \\
    \bottomrule
    \end{tabular}
    }
    \caption{Results on RepliQA and SciQAG. \ul{Underline} indicates improvement of MTP/MPP/NLKG over NTP \textbf{(H1)}. \textit{Italics} indicates improvement of MPP/NLKG over MTP \textbf{(H2)}. \textbf{Bold} indicates the overall best result per model and finetuning method. *: catast. forgetting.}
    \label{tab:results}
\end{table}
\fi

Results are presented in Table~\ref{tab:results}.
Across two tasks, two models, and two fine-tuning modes, instruction-style pretraining tasks outperform the naive NTP baseline 19 out of 24 times. The knowledge-intensive pretraining tasks outperform the MTP baseline in 8 out of 16 cases.
\section{Discussion}

Our results demonstrate support for \textbf{(H1)}: We can successfully improve the performance of instruction-tuned LLMs on a new domain through continual pretraining, a task where past approaches have failed without acccess to the base model.

The support for \textbf{(H2)} is mixed:
Our knowledge-tailored pretraining tasks yield substantial performance gains—often several points—for Llama3.2-3B, but we observe no consistent improvement for Qwen3-1.7B.
This discrepancy is difficult to pinpoint precisely; possible factors include Qwen3’s stronger emphasis on reasoning and its smaller parameter count, which may limit its capacity to benefit from more sophisticated pretraining objectives targeted at knowledge acquisition.

\section{Conclusion}
Addressing the issue of diminishing instruction-following ability during continual pretraining, we introduced a recipe that formulates self-supervised losses in an instruction-response template. 
Our experiments indicate promising accuracy gains with these objectives for question answering tasks.
Future work will explore combining multiple objectives.


\section*{Limitations}
While our proposed framework demonstrates promising results, several limitations remain. First, our current evaluation setup involves training and testing on the full test dataset. Future work will explore model performance when trained on smaller subsets or single-sample scenarios that are typical in test-time training~\citep{sun2020test}.

Second, our scope is limited to question answering tasks. We believe that it is likely that other tasks will also benefit from instruction-style continual pretraining, however, they may require a different set of novel pretraining tasks that are not tailored toward knowledge acquisition. Therefore, we leave a more extensive evaluation to future work.

Third, due to compute constraints we conducted our experiments on relatively small language models of up to 3 billion parameters. It is an open question to what extent our results generalize to larger-scale models.

\section*{Ethical Considerations}
Our experiments focus on a high-resource language, specifically English. While these languages benefit from extensive NLP research and abundant annotated resources, our current evaluation does not extend to low-resource languages such as Faroese or Norwegian. This limitation highlights a potential bias in language coverage. We acknowledge that applying our IKnow approach to low-resource settings could have meaningful implications for linguistic inclusivity. Future work should prioritize expanding the evaluation to such languages to ensure broader applicability and equitable access to language technologies.

\bibliography{custom}

\appendix

\section{Dataset statistics}
\label{app:dataset}
The statistics of dataset is shown in table \ref{tab:dataset}.

\begin{table}[ht]
    \centering
    \resizebox{0.48\textwidth}{!}{
    \begin{tabular}{ccccc}\toprule
      &split&  \# doc&  \# sentence& \# QA pairs\\\midrule
      \ifhalu{
     \href{https://huggingface.co/datasets/pminervini/HaluEval}{HaluEval}&summary&  10k&  340k& -\\}{}
    \href{https://huggingface.co/datasets/ServiceNow/repliqa}{RepliQA}&repliqa\_0&  3k&  160k& 18k\\
    \href{https://github.com/MasterAI-EAM/SciQAG}{SciQAG} &select\_50&  50&  7k& 500\\
    \bottomrule
    \end{tabular}}
    \caption{Statistics of dataset}
    \label{tab:dataset}
\end{table}

     
    

\section{Experimental Setup}
\label{sec:experimental-setup}

We provide the following implementation details to enable better reproducibility and to support the transparency and rigor of our experiments.

\subsection*{Parsing Tools}

\begin{itemize}
    \item \textbf{Dependency Parsing}: We use the \href{https://spacy.io/}{spaCy} library with the \texttt{en\_core\_web\_trf} pretrained transformer-based parsing models.
    
    
    \item \textbf{Constituency Parsing}: We integrate \href{https://github.com/nikitakit/self-attentive-parser}{benepar} with spaCy for constituency parsing, using the pretrained \texttt{benepar\_en3} models.
\end{itemize}

\subsection*{Experimental Framework}

\begin{itemize}
    \item \textbf{Frameworks}:
    \begin{itemize}
        \item \href{https://www.pytorchlightning.ai/}{PyTorch Lightning} is used for modular and scalable training routines.
        \item \href{https://huggingface.co/transformers/}{Hugging Face Transformers} is used for loading and fine-tuning pretrained language models.
    \end{itemize}
\end{itemize}

\subsection*{Training Configuration}

Our final hyper-parameter setting is detailed in table \ref{tab:hyper}.

\begin{table*}[ht]
    \centering
    \resizebox{0.8\textwidth}{!}{
    \begin{tabular}{lll}
    \toprule
         model name&  Llama-3.2-3B-Instruct &Qwen3-1.7B\\
         num\_epochs&  10 &10\\
         batch\_size&  200 &200\\

         weight\_decay & 0 &0\\
         warmup\_steps & 0 &0\\\midrule
         precision & bf16  &bf16\\
         strategy & ddp  &ddp\\
         optimizer&AdamW &AdamW\\
         learning\_rate&  1e-6 - 1e-8 &1e-7 - 1e-9\\
         scheduler& constant &constant\\\midrule
         device & 8*NVIDIA A100-SXM4-80GB&8*NVIDIA A100-SXM4-80GB\\
    \bottomrule
    \end{tabular}
    }
    \caption{Hyper-parmeters for continual pretraining.}
    \label{tab:hyper}
\end{table*}

\begin{itemize}
    \item \textbf{Batch Size}: We use the maximum batch size that can fit into a single GPU.
    \item \textbf{Learning Rate}: 
    We perform experiments on Llama3 with learning rates ranging from 1e-6 to 1e-8 and on Qwen3 with learning rate ranging from 1e-7 to 1e-9 in full fine-tuning and LoRA settings. We search over the range $[5\times10^{-5}, 1\times10^{-9}]$ and select the learning rate casewise with decreasing loss (usually the average loss is from 10 to 1) to balance learning effectiveness and retention of prior knowledge.
    \item \textbf{Epochs}: 
    We train 10 epochs and choose the \textbf{inflection point epoch} (usually 3-7 epochs depending on the dataset and pretraining tasks) for better stability and generalization.
    \item \textbf{Run Count}: Each experiment (training and testing) is conducted \textbf{once}, in accordance with prior work under limited resource conditions.
    \item \textbf{LoRA}: Each task is experimented with the LoRA configuration, see in table \ref{tab:lora}
\end{itemize}

\begin{table}
    \centering
    \begin{tabular}{ll}
    \toprule
         rank& 8\\
         alpha& 16\\
         dropout& 0.05\\
         bias& none\\
 modules&\makecell[l]{q,v,  k, o\\gate, up, down}\\
 \bottomrule
    \end{tabular}
    \caption{LoRA Configuration}
    \label{tab:lora}
\end{table}


\section{Use of Generative AI}
We used ChatGPT and Claude models to assist in writing small functions of our implementations. We also used them to assist in writing by fixing grammar, typos, and general style of our writing. However, we didn't use generative AI for generating ideas or other high-level aspects of our research.

\end{document}